\documentclass[runningheads]{llncs}
\usepackage{amsmath}
\usepackage{amssymb}
\usepackage{xcolor}
\usepackage{booktabs}
\usepackage{algorithm}
\usepackage{algpseudocode}
\usepackage{multirow}
\usepackage{hyperref}
\usepackage[T1]{fontenc}
\usepackage{graphicx}
\graphicspath{ {./Figures/} }

\newcommand{\ds}{\displaystyle}
\newcommand{\Real}{\mathbb{R}}

\newcommand{\xBold}{\boldsymbol{x}}
\newcommand{\xiBold}{\boldsymbol{\xi}}

\newcommand{\aBold}{\boldsymbol{a}}
\newcommand{\bBold}{\boldsymbol{b}}

\begin{document}
\title{Morphological Perceptron with Competitive Layer: Training Using Convex-Concave Procedure}
\titlerunning{MPCL: Training Using Convex-Concave Procedure}
%
\author{Iara Cunha\inst{1}\orcidID{0000-0001-7429-7773} \and
Marcos Eduardo Valle\inst{2}\orcidID{0000-0003-4026-5110}
}
%
%
\institute{Universidade Tecnológica Federal do Paraná (UTFPR), Ponta Grossa, Brazil. \email{iarasilva@utfpr.edu.br} \and
Universidade Estadual de Campinas (UNICAMP), Campinas, Brazil.
\email{valle@ime.unicamp.br}
}
\maketitle              
\begin{abstract}
A morphological perceptron is a multilayer feedforward neural network in which neurons perform elementary operations from mathematical morphology. For multiclass classification tasks, a morphological perceptron with a competitive layer (MPCL) is obtained by integrating a winner-take-all output layer into the standard morphological architecture. The non-differentiability of morphological operators renders gradient-based optimization methods unsuitable for training such networks. Consequently, alternative strategies that do not depend on gradient information are commonly adopted. This paper proposes the use of the convex-concave procedure (CCP) for training MPCL networks. The training problem is formulated as a difference of convex (DC) functions and solved iteratively using CCP, resulting in a sequence of linear programming subproblems. Computational experiments demonstrate the effectiveness of the proposed training method in addressing classification tasks with MPCL networks.

\keywords{Morphological neural network  \and Convex-concave procedure \and Difference of convex optimization.}
\end{abstract}

\section{Introduction}
\label{sec:intro}

The perceptron, introduced by Rosenblatt in 1958, is a fundamental model in neural computation and machine learning. It is characterized as a single linear threshold unit, along with a learning rule that effectively trains the perceptron for linearly separable classification tasks. However, due to the limited computational capabilities of Rosenblatt’s perceptron, the model was successfully expanded into a multi-layer perceptron (MLP) architecture in the 1980s. The success of MLP networks can be attributed not only to their multi-layer structure but also to the efficacy of the backpropagation algorithm. This algorithm enables the computation of gradients used in first-order optimization techniques to minimize the loss function during the training phase.

The morphological perceptron (MP) belongs to the broader category of morphological neural networks, which were developed in the early 1990s by various researchers \cite{Davidson1993,Pessoa1996,Ritter1996,Salembier1992}. Morphological neural networks differ from the traditional perceptron by performing a nonlinear elementary operation based on mathematical morphology before applying the activation function \cite{Sussner2011}. They continue to attract the attention of the scientific community because of their potential to capture geometrical and topological features \cite{Franchi2020DeepNetworks,Groenendijk2023GeometricNetworks,Hu2022LearningSearch,Penaud--Polge2024GroupOperators,Velasco-Forero2022LearnableMorphology}. 

In the MP model, traditional multiplication and addition are substituted with addition and the maximum (or minimum) operations. Using the maximum (or minimum) of sums instead of the sum of products, the computations within the morphological network become non-linear before applying the activation function. This leads to significant geometric and algebraic differences compared to traditional neural network models \cite{Maragos2017,Maragos2021TropicalLearning}. One drawback of morphological neural networks is that training them using backpropagation can be particularly challenging \cite{dimitrova2025hal,Maragos2017}. As a result, many researchers have explored alternative morphological structures or learning rules for these networks \cite{Maragos2017,Franchi2020DeepNetworks,Groenendijk2023GeometricNetworks,Oliveira2021LinearProcedure,Valle2020ReducedClassification}. Inspired by the insightful work of Charisopoulus and Maragos \cite{Maragos2017}, we propose training a morphological perceptron with a competitive layer (MPCL) network using the convex-concave procedure \cite{Lipp2016VariationsProcedure,Yuille2003TheProcedure}. The motivation for considering this model will be explained in the following paragraphs.

Like the traditional perceptron model, a single morphological perceptron has limited computational capabilities, which can be circumvented by employing a multi-layer structure. The multi-layer morphological perceptron (MLMP) network can theoretically solve any binary classification task \cite{Ritter2003,Sussner1998}. Specifically, an MLMP can partition \(\mathbb{R}^n\) into the union of hyperboxes. Since any compact set within \(\mathbb{R}^n\) can be enclosed by a union of hyperboxes, it follows that a binary classification problem can be resolved by encapsulating all positive training data into hyperboxes \cite{Ritter2003,RitterBook}. This characteristic results in a highly interpretable machine learning model, where a pattern is classified as a positive class if it falls within one of the model's hyperboxes. Moreover, the multi-layer morphological perceptron can be adapted for multi-class classification tasks by incorporating a winner-take-all output layer \cite{Sussner2011}. Intuitively, the winner-take-all layer enables the hyperboxes to compete in determining the class of an input pattern, resulting in a model referred to as the morphological perceptron with a competitive layer (MPCL). Training an MPCL network is conducted using the learning rule of an MLMP model along with a one-against-all strategy \cite{Sussner2011}.

The first training algorithms for the MLMPs progressively add hyperboxes, expanding the width of the network sequentially until a stopping criterion is met, such as achieving no misclassifications on the training set. Due to the incremental nature of these algorithms, we refer to them as greedy training algorithms for MLMPs. Examples include the algorithms detailed in \cite{Ritter2003,Sussner1998,Sussner2011}. In contrast to the backpropagation-like methods widely used to train traditional neural networks, the greedy algorithms for training an MLMP do not rely on gradients. While these algorithms train quickly, they often lead to large MLMP models with an excessive number of hyperboxes, making them difficult to interpret. In this paper, we apply the convex-concave procedure to train a fixed-architecture MPCL network, enhancing the interpretability of the resulting machine-learning model. Specifically, in our previous conference paper \cite{CunhaValle2023}, we framed the learning process of a binary MLMP classifier as a difference of convex optimization problem. This was solved using the disciplined convex-concave programming framework, which is available as an extension of \texttt{CVXPY} library for \texttt{Python} \cite{shen2016disciplined}. 
In this paper, we provide a detailed explanation of the convex-concave procedure for training an MLMP network. Additionally, we introduce a regularization term to limit the size of the hyperboxes and propose an initialization method based on k-means++. Finally, based on \cite{Sussner2011}, we extend the model to address multi-class classification problems by considering a morphological perceptron with a competitive layer.

This paper is organized as follows: Section \ref{sec:MP} reviews the MPCL network. In Section \ref{sec:formulation}, we present an optimization problem for training this model. Section \ref{sec:CCP} discusses the application of the convex-concave procedure to address the non-convex optimization problem outlined in Section \ref{sec:formulation}. We provide the results of computational experiments comparing the performance of the proposed training algorithm with similar algorithms from the literature in Section \ref{sec:experiments}. Finally, Section \ref{sec:concluding} offers some concluding remarks.

\section{Morphological Perceptron with Competitive Layer}
\label{sec:MP}

The computations of a morphological perceptron (MP) rely on the algebraic structure $(\bar{\mathbb{R}},\vee,\wedge,+,+')$ from minimax algebra \cite{CuninghameGreen79}, where $\bar{\mathbb{R}}=\{-\infty,+\infty\}$ and $\vee$ and $\wedge$ denote the supremum and infimum operations, respectively. 
The operations ``$+$'' and ``$+'$'' coincide with the real-number addition for finite values, but differ in their treatment of infinities, as specified by 
$(+\infty) + (-\infty) = (-\infty) + (+\infty) = -\infty$ and 
$(+\infty) +' (-\infty) = (-\infty) +' (+\infty) = +\infty$. The conjugate of an element $x \in \bar{\mathbb{R}}$ is defined by 
\begin{equation}
    \label{eq:conjugate}
    \mathtt{conjugate}(x) = \begin{cases}
        -x,& x \in \mathbb{R},\\
        -\infty, & x = +\infty, \\
        +\infty, & x= -\infty.
    \end{cases}
\end{equation}
In this paper, we will focus solely on finite numbers. This allows us to treat $+$ and $+'$ equally (we will only use the symbol $+$) and substitute $\mathtt{conjugate}(x)$ with $-x$. 

\begin{figure} 
\centering
\includegraphics[width=0.7\textwidth]{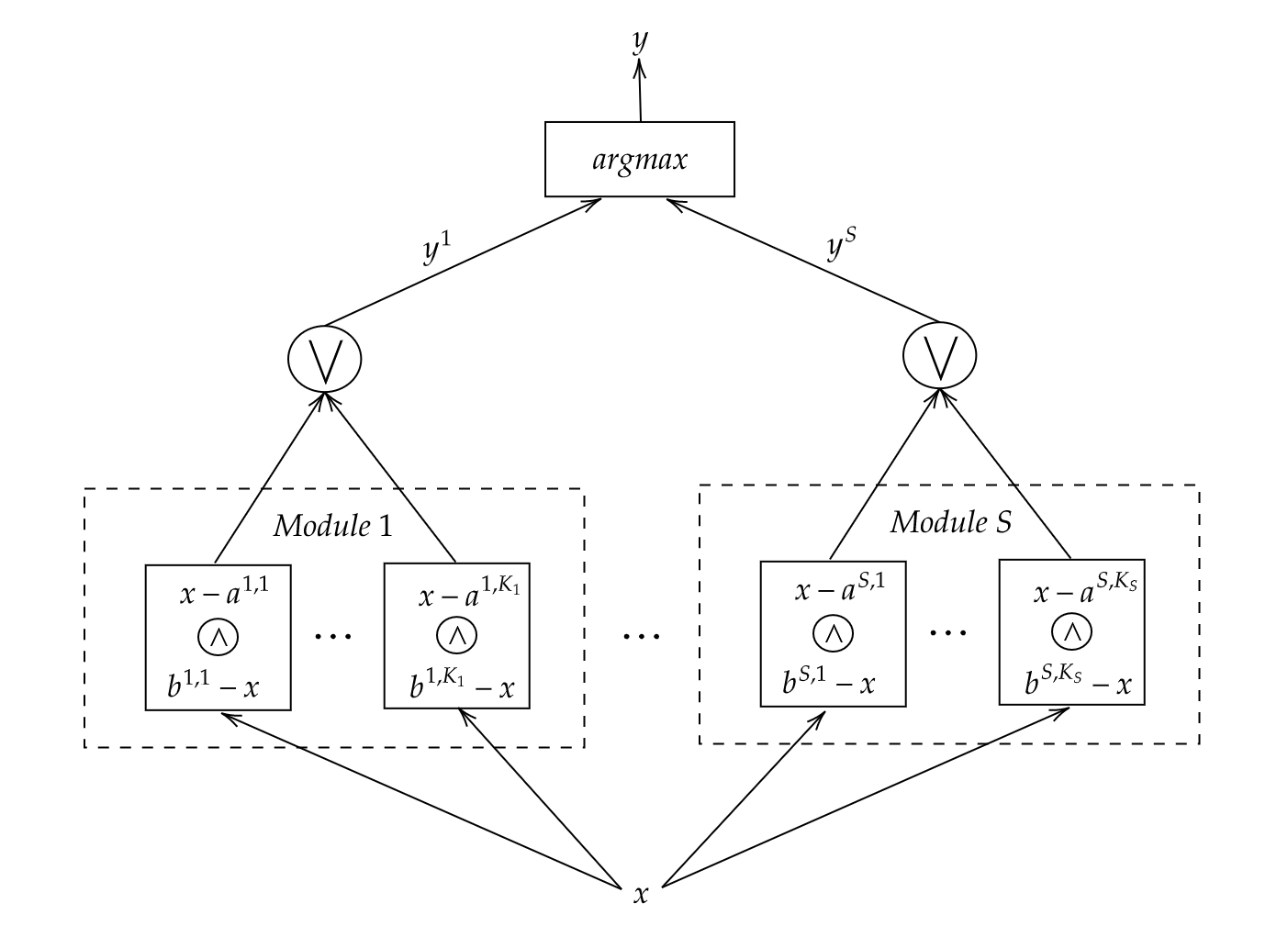}
\caption{Architecture of MPCL classifier.}
\label{fig:ArchitectureCL}
\end{figure}

The model considered in this work is a morphological perceptron with a competitive layer (MPCL), as proposed by Sussner and Esmi \cite{Sussner2011}. Figure \ref{fig:ArchitectureCL} illustrates the architecture of this morphological neural network. An MPCL is structured with the following components: a first layer composed of morphological blocks, a hidden layer containing \( S \) modules, and a winner-takes-all layer represented by an $\text{argmax}$ operation.

The first layer of this neural network consists of blocks, each containing two morphological perceptrons (MPs) that use the identity function as their activation function. One MP performs an erosion, while the other MP implements an anti-dilation, both representing elementary operations from mathematical morphology \cite{Ritter1996,ritter1998}. 
In mathematical terms, let \(\boldsymbol{a}^k = (a_1^k, \ldots, a_n^k) \in \mathbb{R}^n\) and \(\boldsymbol{b}^k = (b_1^k, \ldots, b_n^k) \in \mathbb{R}^n\) represent the synaptic weights of the two MPs in the \(k\)-th block, respectively. The output of the \(k\)-th neuron is given by 
\begin{equation}
\label{eq:neuron_out}
h_k(\xBold) = \left(\bigwedge_{i=1}^n (x_i - a_i^{k})\right) \wedge \left(\bigwedge_{i=1}^n (b_i^{k} - x_i)\right),
\end{equation}
where \(\xBold = (x_1, x_2, \ldots, x_n) \in \mathbb{R}^n\) is the input vector. Note that the morphological block represented by \eqref{eq:neuron_out} characterizes a hyperbox in \(\mathbb{R}^n\). Specifically, we have that \(h_k(\xBold) \geq 0\) if and only if \(a_i^k \leq x_i \leq b_i^k\) for all \(i = 1, \ldots, n\). Equivalently, we have $h_k(\xBold) \geq 0$ if and only if $\xBold \in [\aBold^k,\bBold^k]$. Geometrically, the hypersurface of the function \( h_k \) is a hyper-pyramid with the hyperbox \([\aBold^k, \bBold^k]\) as base.

The hidden layer of an MPCL is composed of $S$ modules. The $s$th module computes the maximum of $K_s$ morphological blocks. Precisely, the output of the $s$th module in the hidden layer is $y^s$ given by 
\begin{equation}
    \label{eq:module-output}
    y^s = \bigvee_{k=1}^{K_s} h_{s,k}(\xBold) = \bigvee_{k=1}^{K_s}  \left ( \bigwedge_{i=1}^n ( x_i-a_i^{s,k}  ) \wedge  \bigwedge_{i=1}^n  (-x_i + b_i^{s,k} ) \right ), \quad \forall \xBold \in \mathbb{R}^n,
\end{equation}
where the index pair ``$s,k$'' in the morphological block means ``$k+\sum_{i=1}^{s-1}K_i$'', for $s=1,\ldots,S$. From a geometrical point of view, the output of the $s$th module characterizes the union of hyperboxes. Accordingly, we have $y^s \geq 0$ if and only if $\xBold \in [\aBold^{s,k},\bBold^{s,k}]$ for some $k \in \{1,\ldots,K_s\}$. 

Finally, the neurons in the output layer compete, and the MPCL assigns the class label corresponding to the largest output from the modules. Formally, the input vector \(\xBold\) is classified as belonging to a class \(s^*\) such that $y^{s^*} = \max \{y^1, \ldots, y^S\}$. Since the probability of having two boxes with the same value is nearly zero, we can take \(s^* = \arg\max \{y^1, \ldots, y^S\}\) as the first index that yields the maximum. Thus, the MPCL assigns the class \(s\) associated with the hyperbox \([\mathbf{a}^{s,k}, \mathbf{b}^{s,k}]\) that best encloses the input \(\xBold\). Based on this remark, Sussner and Esmi proposed an algorithm that recursively constructs hyperboxes containing only data from one class \cite{Sussner2011}. Alternative constructive morphological perceptron algorithms are discussed in \cite{Ritter2003,Sussner1998}. From an optimization standpoint, these constructive algorithms utilize greedy heuristics to reduce classification error. The following section presents an alternative approach for training an MPCL network.



\section{Optimization Problem for Training the MPCL Network}
\label{sec:formulation}

This section presents a constrained convex-concave optimization problem for training the MPCL network to solve multi-class classification tasks. Accordingly, consider a training set
\begin{equation}
\mathcal{T} = \{( \boldsymbol{x}^{(j)},y^{(j)}):j=1,\ldots,M \} \subseteq \mathbb{R}^n \times \mathcal{S}
\label{eq:TrainingSet}
\end{equation}
where $M$ is the number of samples, $\boldsymbol{x}^{(j)} = (x_1^{(j)}, x_2^{(j)}, \ldots, x_n^{(j)}) \in \Real^n$ is the $j$th feature vector, and $y^{(j)} \in \mathcal{S}$ is its label. We will identify the label set with $\mathcal{S}=\{1,\ldots,S\}$ for simplicity. 

The MPCL training aims to create hyperboxes that efficiently enclose the training data from a single class. The goal is accomplished using a one-against-all strategy. Precisely, let us define the sets of negative and positive samples with respect to a class label $s \in \mathcal{S}$ as follows, respectively:
\begin{equation} 
\label{eq:sets}
C_0^s = \{\boldsymbol{x}^{(j)}:y^{(j)} \neq s \} \quad \text{and} \quad C_1^s = \{\boldsymbol{x}^{(j)}:y^{(j)} = s \}.
\end{equation}
The training process aims to find a family of hyperboxes 
\begin{equation} 
\label{eq:Hyperbox_set}
\mathcal{H}_s= \left \{[\aBold^{s,1},\bBold^{s,1}], [\aBold^{s,2},\bBold^{s,2}],\ldots,[\aBold^{s,K_s},\bBold^{s,K_s}] \right\},
\end{equation}
whose union contains samples from class \( s \) while excluding patterns from other classes.  
%
%
%
%
%
%
In mathematical terms, the hyperboxes are obtained by solving the optimization problem:
\begin{equation}
\label{eq:InitialModel}
\begin{array}{cl} 
\ds \min_{\aBold^s, \bBold^s, \xiBold^s} \quad & \ds \sum_{l = 1}^M \xi_l^s  
+ \gamma \sum_{k=1}^{K_s} \sum_{i=1}^n (b^{s,k}_i-a^{s,k}_i),\\
\textrm{s.t.} & \ds \bigvee_{k=1}^{K_s}  \left ( \bigwedge_{i=1}^n ( x_i^{(l)}-a_i^{s,k}  ) \wedge  \bigwedge_{i=1}^n  (-x_i^{(l)} + b_i^{s,k} ) \right )  \leq \xi_l^s \quad \mbox{ if } \ \xBold^{(l)} \in C_0^s, \\ 
& \ds \bigvee_{k=1}^{K_s}  \left ( \bigwedge_{i=1}^n ( x_i^{(l)}-a_i^{s,k}  ) \wedge  \bigwedge_{i=1}^n  (-x_i^{(l)} + b_i^{s,k} ) \right )  \geq - \xi_l^s \ \mbox{ if }  \xBold^{(l)}  \in C_1^s, \\
& \aBold^s \leq \bBold^s, \quad  \boldsymbol{0} \leq \xiBold^s, 
\end{array}
\end{equation}
where \(\aBold^{s} = [\aBold^1, \ldots, \aBold^{K_s}]\) and \(\bBold^{s} = [\bBold^1, \ldots, \bBold^{K_s}]\) are the lists of lower and upper vertices of the hyperboxes, \(\xiBold^s = (\xi_1^s, \ldots, \xi_M^s) \in \mathbb{R}^M\) is the vector of slack variables, and \(\gamma\) is a regularization parameter. 

The optimization problem presented in \eqref{eq:InitialModel} is based on the approach by Charisopoulos and Maragos \cite{Maragos2017}, who utilize a formulation similar to support vector machines but employ a minimax algebraic structure. The first two constraints address classification errors for negative and positive samples, respectively. Specifically, a sample \(\xBold^{(l)}\) that does not belong to class \(s\) should not be included in the union of the hyperboxes. Mathematically, this is expressed using \eqref{eq:module-output} as the inequality \(\bigvee_{k=1}^{K_s} h_{s,k}(\xBold^{(l)}) < 0\). By incorporating the slack variable \(\xi_l^s\) on the right-hand side, we allow for the possibility of violating this inequality, resulting in the first constraint in \eqref{eq:InitialModel}. Conversely, if the sample \(\xBold^{(l)}\) belongs to class \(s\), the inequality \(\bigvee_{k=1}^{K_s} h_{s,k}(\xBold^{(l)}) > 0\) should hold. Including the slack variable \(\xi_l^s\) on the right-hand side of this inequality leads to the second constraint in \eqref{eq:InitialModel}. The final two inequality constraints ensure that \(\aBold^s\) and \(\bBold^s\) define valid hyperboxes and that the slack variables remain non-negative. It is important to note that \(\xi_l^s>0\) accounts for a misclassified sample \(\xBold^{(l)}\). Consequently, the first term in the objective function aims to minimize the slack variables. The second term, which can be interpreted as a regularization term, aims to minimize the size of the hyperboxes.



Although the objective of the optimization problem \eqref{eq:InitialModel} is linear, the constraints are neither convex nor concave functions, making it a challenging optimization problem. Nevertheless, it is possible to reformulate the constraints of \eqref{eq:InitialModel} using the difference of convex (DC) functions \cite{CunhaValle2023}. Recall that a real-valued function $\phi$ is called a difference of convex (DC) function if there exist two convex functions $f$ and $g$ such that $\phi = f-g$. A DC optimization problem has the form
\begin{equation}
    \label{eq:DC-Problem}
    \begin{array}{rl}
       \displaystyle{\min_{\boldsymbol{v}}}  & f_0(\boldsymbol{v})-g_0(\boldsymbol{v})  \\
       \textrm{s.t.} & f_i(\boldsymbol{v})-g_i(\boldsymbol{v}) \leq 0, \quad \forall i = 1,\ldots,N_C,
    \end{array}
\end{equation}
where $\boldsymbol{v}$ is the optimization variable and $f_i,g_i$ are convex functions for all $i \in \{0,1,\ldots,N_C\}$, with $N_C$ being the number of constraints \cite{Lipp2016VariationsProcedure}.

The function $h_k$ given by \eqref{eq:neuron_out}, which computes the minimum of linear functions, is concave. Hence, the function $\Psi(\cdot,\aBold^{s,k},\bBold^{s,k}) = -h_k$ given by the following expression in terms of the input $\xBold \in \mathbb{R}^n$ and the parameters $\aBold^{s,k} \in \mathbb{R}^n$ and $\bBold^{s,k} \in \mathbb{R}^n$ is convex:
\begin{equation}
\label{eq:psiFunction}
\Psi(\xBold,\aBold^{s,k},\bBold^{s,k})=  \left(\bigvee_{i=1}^n (a_i^{s,k}-x_i)\right) \vee \left(\bigvee_{i=1}^n (x_i - b_i^{s,k})\right) 
\equiv \bigvee \begin{bmatrix} \aBold^{s,k}-\xBold \\ \xBold - \bBold^{s,k} \end{bmatrix}.
\end{equation}
Moreover, we can write the output of a module given by \eqref{eq:module-output} as a function $\varphi$, which depends on the input $\xBold \in \mathbb{R}^n$ and the lists \(\aBold^s = [\aBold^{s,1}, \ldots, \aBold^{s,K_s}]\) and \(\bBold^s = [\bBold^{s,1}, \ldots, \bBold^{s,K_s}]\) of the lower and upper vertices of the hyperboxes, using the equation:
\begin{equation}
\label{eq:phiFunctionR}
\varphi(\xBold,\aBold^s,\bBold^s) = \ds \bigvee_{k=1}^{K_s} \left ( \bigwedge_{i=1}^n ( x_i -a_i^{s,k}) \wedge \bigwedge_{i=1}^n (  b_i^{s,k}-x_i)  \right )
= \ds \bigvee_{k=1}^{K_s}  \left ( - \Psi(\xBold, \aBold^{s,k},\bBold^{s,k})\right).
\end{equation}
From the proof of Proposition 3.1 in \cite{Bagirov:2020}, the function $\varphi$ given by \eqref{eq:phiFunctionR} can be expressed as a DC function:
\begin{equation}
\varphi(\xBold, \aBold^{s},\bBold^{s}) = f(\xBold, \aBold^{s},\bBold^{s})-g(\xBold, \aBold^{s},\bBold^{s})
\label{eq:DCfunction}
\end{equation}
where 
\begin{equation}
\label{eq:f}
f (\xBold, \aBold^{s},\bBold^{s}) = \ds \bigvee_{k=1}^{K_s}  \left\{ \sum_{j \neq k}^{K_s} \Psi(\xBold, \aBold^{s,j}, \bBold^{s,j}) \right\}, 
\end{equation}
and
\begin{equation}
\label{eq:g}
g (\xBold, \aBold^{s},\bBold^{s}) = \ds \sum_{k=1}^{K_s} \Psi(\xBold, \aBold^{s,k}, \bBold^{s,k}).
\end{equation}
%
%
%
%
%
%
%
Hence, the optimization problem \eqref{eq:InitialModel} can be rewritten as the DC problem \cite{Lipp2016VariationsProcedure,shen2016disciplined}:
\begin{equation}
\begin{array}{rlclll} 
\ds \min_{\aBold^s, \bBold^s, \xiBold^s} &  \multicolumn{3}{l}{\quad \ds \sum_{l = 1}^M \xi_l^s + \gamma \sum_{k=1}^{K_s} \sum_{i=1}^n (b^{s,k}_i-a^{s,k}_i) }  \\
\textrm{s.t.} & f(\xBold^{(l)},\aBold^s, \bBold^s) -g(\xBold^{(l)},\aBold^s, \bBold^s) & -\xi_l^s & \leq 0 & \mbox{ if }  \xBold^{(l)} \in C_0^s\\
&g(\xBold^{(l)},\aBold^s, \bBold^s) -f(\xBold^{(l)},\aBold^s, \bBold^s) & -\xi_l^s  & \leq 0 &   \mbox{ if }  \xBold^{(l)}  \in C_1^s \\
&\aBold^s - \bBold^s \leq \mathbf{0},  \quad \xiBold^s \geq \mathbf{0}. &  &  &
\end{array}
\label{eq:mainModel_DC}
\end{equation}
The optimization problem presented in \eqref{eq:mainModel_DC}, which is equivalent to problem \eqref{eq:InitialModel}, can be solved using the Disciplined Convex-Concave Programming (DCCP) extension of the \texttt{CVXPY} library in \texttt{Python} \cite{Diamond2016cvxpy}, as demonstrated in \cite{CunhaValle2023}. In the following section, we eliminate the need for the DCCP framework by detailing the application of the convex-concave procedure (CCP) to solve \eqref{eq:mainModel_DC}.

\section{Convex-Concave Procedure}
\label{sec:CCP}

In this section, we outline the convex-concave procedure (CCP) for solving the optimization problems \eqref{eq:mainModel_DC} that incorporate both convex and concave components in the constraints \cite{Lipp2016VariationsProcedure,Yuille2003TheProcedure}. The CCP is a powerful heuristic method used to find local solutions of DC optimization problems in the form \eqref{eq:DC-Problem}. The basic idea of the CCP is to replace the convex function $g_i$ (or the concave function $-g_i$) by a linear approximation. Accordingly, recall that $\partial g_i(\boldsymbol{v}_t)$ is a subgradient of $g_i$ at $\boldsymbol{v}_t$ if 
\begin{equation}
\label{eq:subgradient}
    g_i(\boldsymbol{v}) \geq g_i(\boldsymbol{v}_t)+\langle \partial g_i(\boldsymbol{v}_t),\boldsymbol{v}-\boldsymbol{v}_t \rangle := \mathcal{L}_{g_i,\boldsymbol{v}_t}(\boldsymbol{v}),
\end{equation}
where $\mathcal{L}_{g_i,\boldsymbol{v}_t}$ denotes a linear approximation of $g_i$ on $\boldsymbol{v}_t$. Given a feasible initial point $\boldsymbol{v}_0$, the CCP computes $\boldsymbol{v}_{t+1}$ by solving iteratively the convex subproblem
\begin{equation}
    \label{eq:CCP-SubProblem}
    \begin{array}{rl}
       \displaystyle{\boldsymbol{v}_{t+1} = \mathop{\textrm{argmin}}_{\boldsymbol{v}}}  & f_0(\boldsymbol{v})-\mathcal{L}_{g_0,\boldsymbol{v}_t}(\boldsymbol{v})  \\
       \textrm{s.t.} & f_i(\boldsymbol{v})-\mathcal{L}_{g_i,\boldsymbol{v}_t}(\boldsymbol{v}) \leq 0, \quad \forall i=1,\ldots,N_C,
    \end{array}
\end{equation}
until a stopping criterion is satisfied. The following presents the convex subproblems derived from applying the CCP to the DC optimization problem \eqref{eq:mainModel_DC}. 

The objective function of the DC optimization problem \eqref{eq:mainModel_DC} is linear, so there is no need to linearize it. Similarly, the constraints $\aBold^s - \bBold^s \leq \mathbf{0}$ and $\sigma^s \geq \mathbf{0}$ also do not require linearization. Therefore, we only need to focus on simplifying or linearizing the first two inequality constraints. 

First of all, recall that $f(\xBold^{(l)},\aBold^s, \bBold^s) -g(\xBold^{(l)},\aBold^s, \bBold^s) = \bigvee_{k=1}^{K_s}(-\Psi(\xBold,\aBold^{s,k},\bBold^{s,k})$. Thus, the first inequality constraint of \eqref{eq:mainModel_DC} can be expressed as
\begin{equation}
 \bigvee_{k=1}^{K_s}(-\Psi(\xBold,\aBold^{s,k},\bBold^{s,k}) - \xi_l^s \leq 0, \quad \forall \xBold^{(l)} \in C_0^s.   
\end{equation}
Since the maximum is the greatest element, this inequality holds if and only if
\begin{equation}
    - \xi_l^s  -\Psi(\xBold^{(l)},\aBold^{s,k}, \bBold^{s,k})  \leq  0,  \quad \forall k = \{1, \ldots, K_s\} \quad \text{and}\quad  \forall \xBold^{(l)} \in C_0^s,
\end{equation}
where $\Psi$ is the convex function defined by \eqref{eq:psiFunction}. Given a training feature vector $\xBold^{(l)} \in \Real^n$, we can compute a linear approximation of $\Psi$ on $(\aBold^{s,k}_t,\bBold^{s,k}_t)$, where $\aBold^{s,k}_t$ and $\bBold^{s,k}_t$ denote the lower and upper vertexes of the $k$th hyperbox of module $s$ at iteration $t$ of the CCP algorithm.
Using the concept of subgradient, we conclude that a linear approximation of $\Psi$ on $(\aBold^{s,k}_t,\bBold^{s,k}_t)$ is given by
\begin{equation}
    \label{eq:L_Psi}
    \mathcal{L}_{\Psi,\aBold^{s,k}_t,\bBold^{s,k}_t}(\xBold^{(l)},\aBold^{s,k},\bBold^{s,k}) = 
    \begin{cases}
        a_{i^*}^{s,k}-x_{i^*}, & i^* \leq n, \\
        x_{i^*-n}-b_{i^*-n}^{s,k}, & i^* > n,
    \end{cases}
\end{equation}
where the index 
\begin{equation}
    \label{eq:i^*}
    i^* = \arg\max\big\{[\aBold^{s,k}_t-\xBold^{(l)},\xBold^{(l)} - \bBold^{s,k}_t]\big\} \in \{1,\ldots,2n\},
\end{equation}
yields the maximum of $[\aBold^{s,k}_t-\xBold^{(l)},\xBold^{(l)} - \bBold^{s,k}_t] \in \Real^{2n}$. Note that the linear approximation $\mathcal{L}_{\Psi,\aBold^{s,k},\bBold^{s,k}}$ depends on $\aBold^{s,k}_t$ and $\bBold^{s,k}_t$, as well as on $\xBold^{(l)}$, through the index $i^*$ which depends on $s,k,t$, and $l$, that is, $i^* \equiv i^*(s,k,t,l)$. Therefore, the first set of inequalities in \eqref{eq:mainModel_DC} leads to the following constraints for the CCP optimization subproblem for all $\xBold^{(l)} \in C_0^s$ and $k=1,\ldots,K_s$ with $i^*$ given by \eqref{eq:i^*} for each $s$, $k$, $t$, and $l$:
\begin{equation}
\begin{cases}
        -a_{i^*}^{s,k} - \xi_l^s \leq -x_{i^*}^{(l)}, & \text{if } i^* \leq n, \\
        b_{i^*-n}^{s,k} - \xi_l^s \leq x_{i^*-n}^{(l)}, & \text{if } i^* > n.
    \end{cases}
\end{equation}

In a similar fashion, a linear approximation of the convex function $f$ given by \eqref{eq:f} on $(\aBold^s_t,\bBold^s_t)$ for a fixed feature vector $\xBold^{(l)}$ can be expressed in terms of the linear approximations of $\Psi$ employing the equation
\begin{equation}
    \label{eq:L_f}
    \mathcal{L}_{f,\aBold^{s}_t,\bBold^{s}_t}(\xBold^{(l)},\aBold^s,\bBold^s) = 
    \sum_{j \neq k^*} \mathcal{L}_{\Psi,\aBold^{s,j}_t,\bBold^{s,j}_t}(\xBold^{(l)},\aBold^{s,j},\bBold^{s,j}),
\end{equation}
where $k^* \equiv k^*(s,t,l)$ is defined by
\begin{equation}
    \label{eq:k^*}
    k^* = \mathop{\arg\max}_{k=1:K_s} \sum_{j \neq k} \Psi(\xBold^{(l)},\aBold_t^{s,j},\bBold_t^{s,j}).
\end{equation}
As a consequence, the second set of inequalities in \eqref{eq:mainModel_DC} leads to the following constraints for the CCP optimization subproblem for all $\xBold^{(l)} \in C_1^s$ with $k^*$ given by \eqref{eq:k^*} and for all $i=1,\ldots,2n$:
\begin{equation}
\begin{cases}
        a_{i}^{s,k^*} - \xi_l^s \leq x_{i}^{(l)}, & \text{if } i \leq n, \\
        -b_{i-n}^{s,k^*} - \xi_l^s \leq -x_{i-n}^{(l)}, & \text{if } i > n.
    \end{cases}
\end{equation}

Concluding, the CCP applied to solve the DC optimization problem \eqref{eq:mainModel_DC} results in solving iteratively the following sequence of linear programming problems for training the MPCL:
\begin{equation}
    \label{eq:DClinear}
    \begin{array}{rl}
       \displaystyle{\boldsymbol{a}^s_{t+1},\bBold^s_{t+1} = \mathop{\textrm{argmin}}_{\boldsymbol{a}^s,\bBold^s,\xiBold^s}}  & \displaystyle{\sum_{l = 1}^M \xi_l^s + \gamma \sum_{k=1}^{K_s} \sum_{i=1}^n (b^{s,k}_i-a^{s,k}_i)}  \\
       \textrm{s.t.} & 
       \begin{cases}
        -a_{i^*}^{s,k} - \xi_l^s \leq -x_{i^*}^{(l)}, & \text{if } i^* \leq n, \\
        b_{i^*-n}^{s,k} - \xi_l^s \leq x_{i^*-n}^{(l)}, & \text{if } i^* > n,
    \end{cases} \quad \begin{array}{l}
         \forall k=1,\ldots,K_s,  \\
         \forall \xBold^{(l)} \in C_0^s,
    \end{array} \\
    & \begin{cases}
        a_{i}^{s,k^*} - \xi_l^s \leq x_{i}^{(l)}, \quad \; & \text{if } i \leq n, \\
        -b_{i-n}^{s,k^*} - \xi_l^s \leq x_{i-n}^{(l)}, & \text{if } i > n,
    \end{cases} \quad \begin{array}{l}
    \forall i=1,\ldots,2n,  \\
    \forall \xBold^{(l)} \in C_1^s,
    \end{array} \\
    & \aBold^s - \bBold^s \leq \mathbf{0},  \quad \xiBold^s \geq \mathbf{0},
    \end{array}
\end{equation}
where $k^* \equiv k^*(s,t,l)$ and $i^* \equiv i^*(s,k^*,t,l)$ given respectively by \eqref{eq:k^*} and \eqref{eq:i^*}. 
The linear optimization subproblems are solved iteratively until there is no significant difference in the objective function, or a maximum number of iterations is reached. Also, the sequence of problems is solved for each class label \( s = 1, \ldots, S \).

We would like to clarify that we initialized the hyperboxes using the K-means++ algorithm \cite{arthur2007}. K-means++ is employed to select initial centroids for K-means clustering. This algorithm ensures that the selected centroids are well-distributed, which helps minimize the risk of poor initialization. By utilizing K-means++, the hyperboxes are initialized with $\aBold^{s,k}=\bBold^{s,k} = \boldsymbol{c}^{s,k}$, where $\boldsymbol{c}^{s,k}$ represents the $k$th centroid determined by the K-means++ algorithm using the dataset $C_1^s$, for $s=1,\ldots,S$.

\section{Computational Experiments}
\label{sec:experiments}

In this section, we assess the classification accuracy and computational efficiency of four morphological perceptron-based classifiers: the proposed MPCL-CCP, the MPCL-DCCP method from our previous work \cite{CunhaValle2023}, the MPCL-Greedy approach based on Sussner and Esmi \cite{Sussner2011}, and MPCL-Adam, trained using a backpropagation-like technique based on the Adam optimizer \cite{kingma2015adam}.

All methods were implemented in Python 3.9, utilizing the CVXPY optimization library with the CLARABEL and MOSEK solvers, as well as TensorFlow for the MPCL-Adam approach. A one-against-all strategy was employed for MPCL-Adam training, as in the other methods. As commonly done in binary classification neural networks, the hyperboxes of the $s$th module, defined by \eqref{eq:module-output}, are determined by minimizing the binary cross-entropy of the logits. The default parameters of Keras, version 3.10, were used for the Adam optimizer, including a learning rate of 0.001, and the number of training epochs was fixed at 100. To reduce sensitivity to initialization \cite{dimitrova2025hal}, the hyperbox parameters were initialized using the K-means++ technique, as described in Section \ref{sec:CCP}.

\begin{figure}[t]
    \centering
    \begin{tabular}{cc} 
    a) Boxplot of the F1-score in the test set. & b) Diagram of a hypothesis test.\\
    \includegraphics[width=0.55\columnwidth]{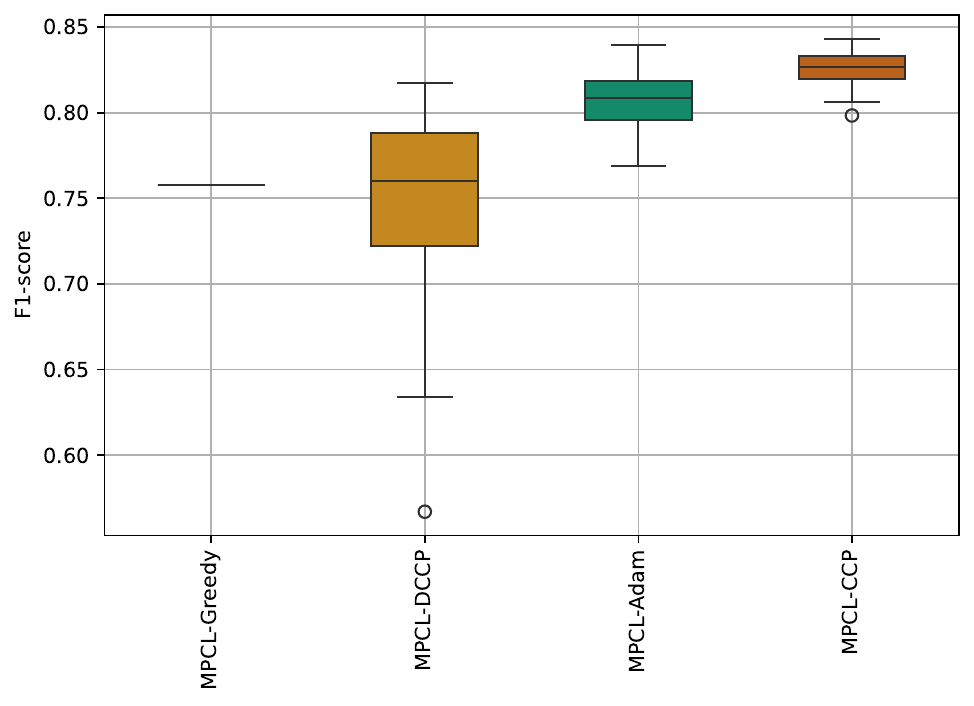} &
    \includegraphics[width=0.45\columnwidth]{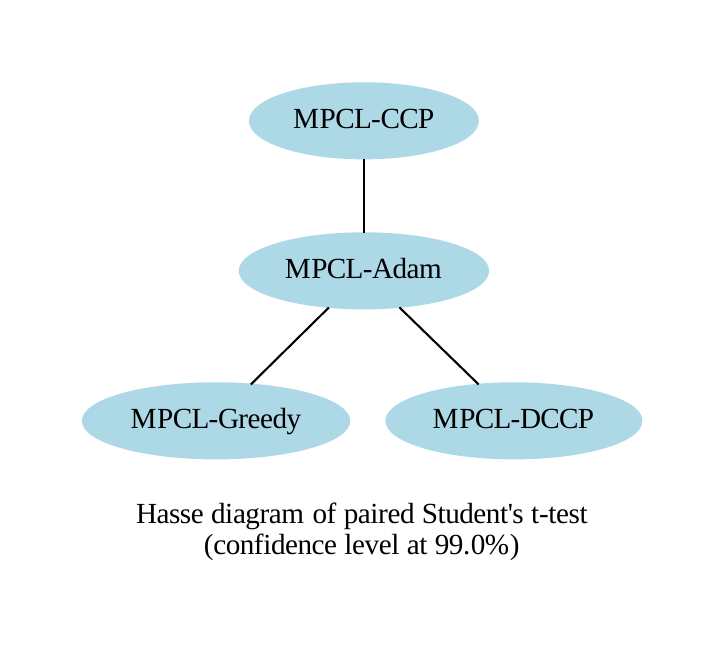}
    \end{tabular}
    \caption{Test set evaluation: F1-score and Hasse diagram of a hypothesis test.}
    \label{fig:boxplot_testF1}
\end{figure}

For MPCL-CCP, MPCL-DCCP, and MPCL-Adam, the architecture includes $K_s = 4$ hyperboxes per class. Experiments were conducted on a synthetic dataset with three classes, generated using the \texttt{make\_blobs} function, containing 1,200 samples with two features, 12 cluster centers, a cluster standard deviation of 1.5, and a fixed random seed of 42. Each method was trained 50 times to account for optimization variability. Performance was measured using the F1-score and misclassification rate. Results are presented in Table \ref{tab:results_blobs} and Figure \ref{fig:boxplot_testF1}.

\begin{table}[t]
    \centering
    \caption{Classification performance on the three-class synthetic dataset.}
    \label{tab:results_blobs}
    \begin{tabular}{lcccc}
        \hline
        \multirow{2}{*}{Method} & \multicolumn{2}{c}{F1-score} & \multicolumn{2}{c}{Misclassified (\%)} \\
        \cline{2-5}
        & Train & Test & Train & Test \\
        \hline
        MPCL-Greedy  & \textbf{0.88}  & 0.76 & \textbf{11.6}   & 24.2  \\
        MPCL-DCCP    & $0.75_{\pm0.06}$  & $0.75_{\pm0.06}$ & $23.7_{\pm5.5}$   & $25.3_{\pm5.3}$  \\
        MPCL-Adam    & $0.81_{\pm0.01}$  & $0.81_{\pm0.02}$  & $19.1_{\pm1.4}$     & $19.3_{\pm1.7}$\\
        MPCL-CCP     & $0.82_{\pm0.01}$  & $\mathbf{0.83_{\pm0.01}}$ & $17.2_{\pm0.6}$   & $\mathbf{17.5_{\pm1.0}}$  \\
        \hline
    \end{tabular}
\end{table}

Note from Table \ref{tab:results_blobs} and Figure \ref{fig:boxplot_testF1} that the MPCL-CCP achieved the best performance, achieving an average F1-score of 0.83 and the lowest error rate of 17.5\% on the test set. Moreover, the MPCL-CCP yielded the lowest variability, reinforcing its robustness and stability. The MPCL-Adam obtained similar F1-scores (0.81) but with higher variability, indicating competitive performance but higher sensitivity. In contrast, MPCL-DCCP showed lower performance (average F1-score 0.75) and high variability, as reflected by the standard deviation and wider spread in the boxplot in Figure \ref{fig:boxplot_testF1}a). MPCL-Greedy, while achieving a high training F1-score of 0.88, experienced a substantial performance drop on the test set, yielding an F1-score of 0.76 and an error rate of 24.2\%. Recall that the MPCL-Greedy method is deterministic and, thus, it does not exhibit any variation in its performance metrics. The Hasse diagram in Figure \ref{fig:boxplot_testF1}b) summarizes the outcome of paired Student's t-tests with a 99\% confidence level, with pairwise winners indicated above, on the test set \cite{weise15}. The performance hierarchy highlights MPCL-CCP as the most accurate approach. 

Regarding computational efficiency, MPCL-CCP completed training in an average of 1.72 seconds, approximately 132 times faster than MPCL-DCCP (228.05 seconds). MPCL-Adam required 40.79 seconds, over 23 times longer than MPCL-CCP, due in part to the overhead of managing the TensorFlow computational graph. The MPCL-Greedy was extremely fast, completing training in 0.05 seconds ($\approx$3.2\% of MPCL-CCP’s time).

\begin{figure}[t]
\centering
\begin{tabular}{cc}
a) MPCL-Greedy  & b) MPCL-DCCP \\
\includegraphics[width=0.45\textwidth]{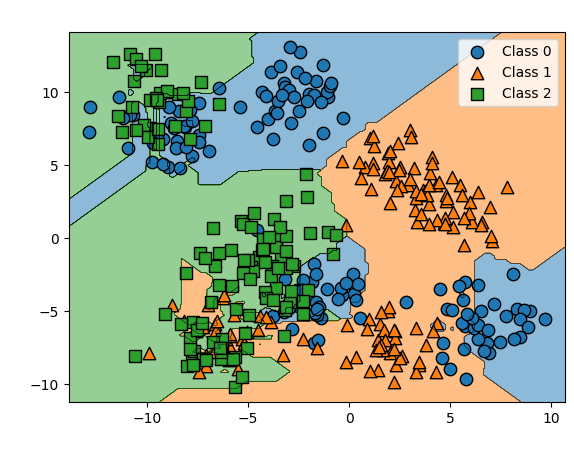}  &
\includegraphics[width=0.45\textwidth]{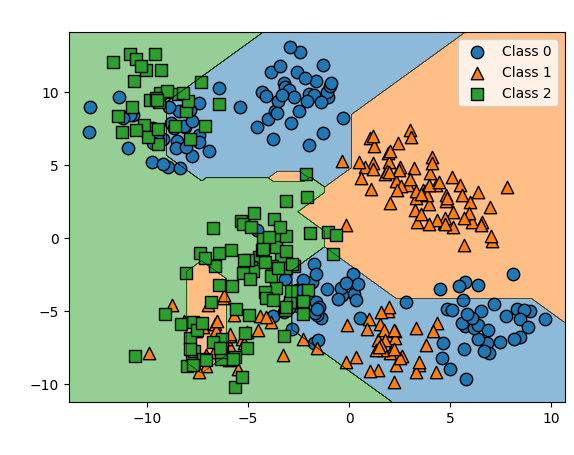} \\ 
c) MPCL-Adam & d) MPCL-CCP \\
\includegraphics[width=0.45\textwidth]{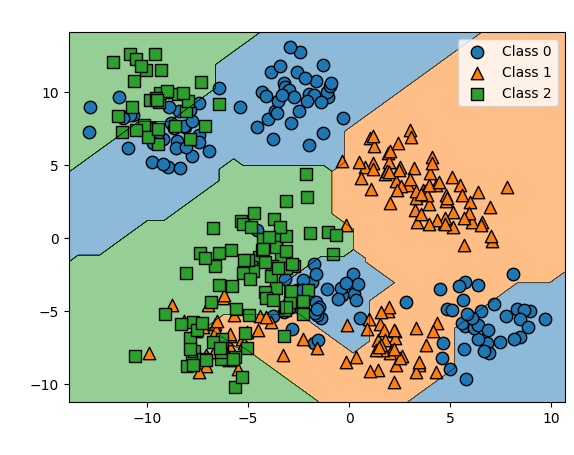}  &
\includegraphics[width=0.45\textwidth]{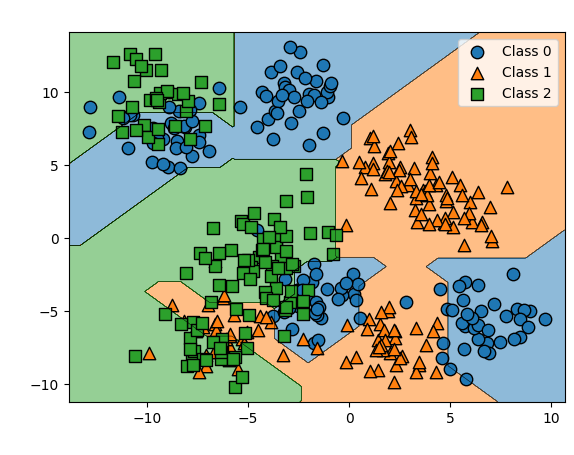} 
\end{tabular}
\caption{Decision surfaces generated by the MPCL classifiers on the test set.}
\label{fig:decision_surface}
\end{figure}

\begin{figure}[t]
\centering
\begin{tabular}{cc}
a) MPCL-Greedy  & b) MPCL-DCCP \\
\includegraphics[width=0.45\textwidth]{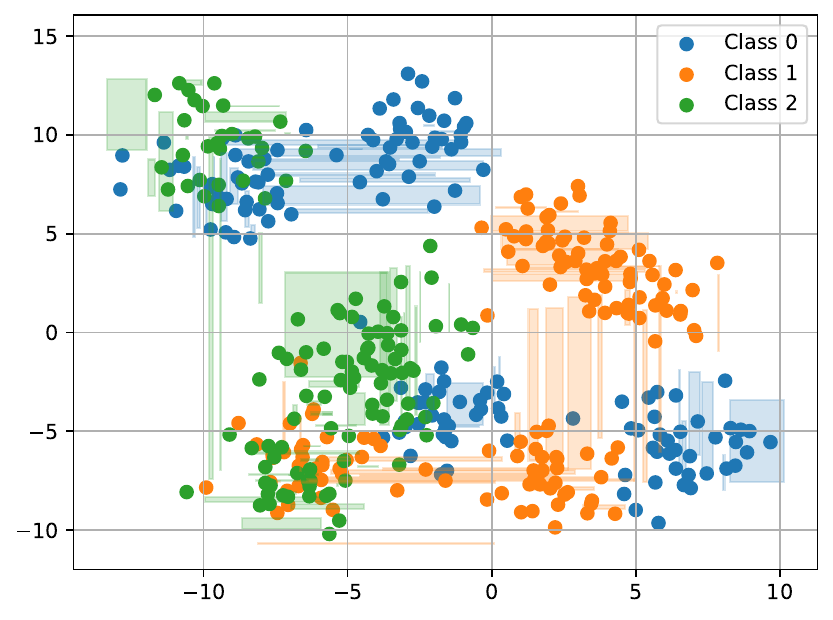}  &
\includegraphics[width=0.45\textwidth]{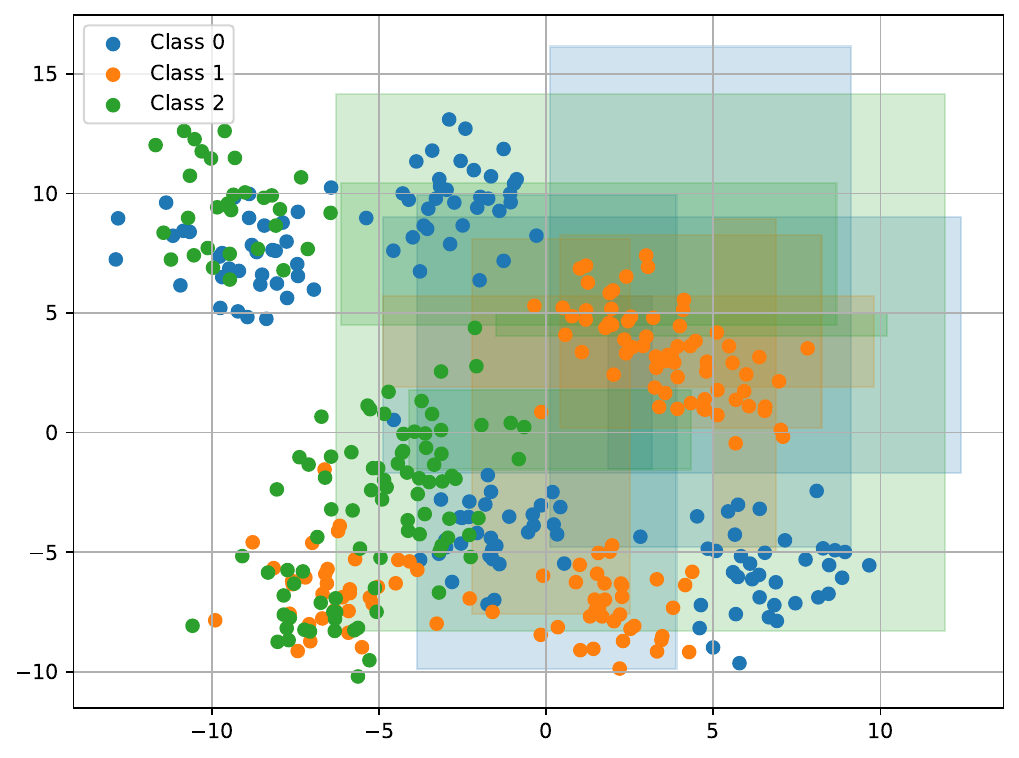} \\ 
c) MPCL-Adam & d) MPCL-CCP \\
\includegraphics[width=0.45\textwidth]{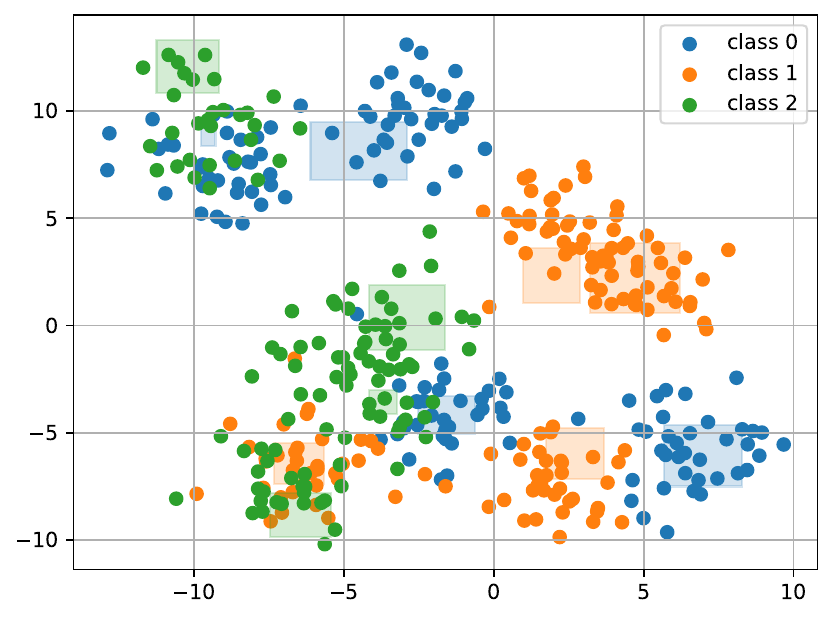} &
\includegraphics[width=0.45\textwidth]{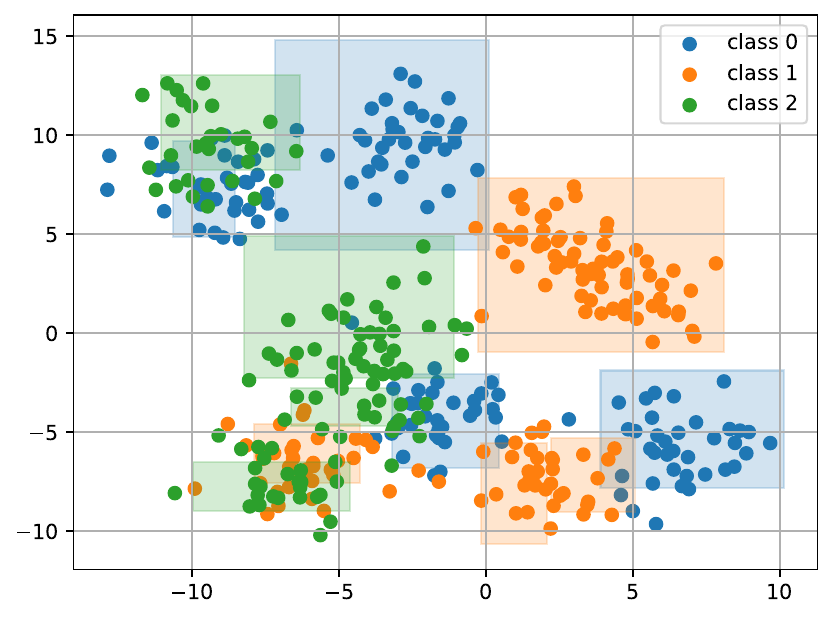} 
\end{tabular}
\caption{Hyperboxes generated by the MPCL classifiers on the test set.}
\label{fig:hyperboxes}
\end{figure}

Finally, Figures \ref{fig:hyperboxes} and \ref{fig:decision_surface} shows the hyperboxes and decision surfaces of the MPCL models. Note from Figures \ref{fig:hyperboxes} the significant differences between the hyperboxes produced by the models. In particular, note that the MPCL-CCP features hyperboxes, which enable high interpretability. The MPCL-Adam’s also features interpretable hyperboxes, but those produced by the MPCL-CCP better fit the training data, which contributes to its superior generalization capability. Finally, the decision surfaces shown in Figure \ref{fig:decision_surface} confirm that MPCL-CCP provides the most accurate class separation.

\section{Concluding Remarks} \label{sec:concluding}

This paper introduced the MPCL-CCP, a morphological perceptron with a competitive layer trained using the convex-concave procedure (CCP). The training features a difference-of-convex (DC) optimization problem whose goal is to minimize misclassification errors, while incorporating a regularization term that reduces the sizes of hyperboxes. The CCP enables efficient training via iterative linear programming without requiring gradients.

Experimental results on a synthetic dataset demonstrate that MPCL-CCP achieves an excellent predictive performance, with high interpretability and low computational cost. Accordingly, the MPCL-CCP yielded the highest test F1-score (0.83) and the lowest error rate (17.5\%) with low variability, outperforming MPCL-Greedy, MPCL-DCCP, and MPCL-Adam. Although MPCL-Adam reached a similar F1-score (0.81), its results were less consistent and computationally more demanding. MPCL-Greedy, while extremely fast and achieving high training scores, showed poor generalization, and MPCL-DCCP exhibited both lower accuracy and higher variability. Qualitative analyses of hyperboxes and decision surfaces confirm that MPCL-CCP produces more interpretable structures that better capture the data distribution than the other approaches.

Overall, MPCL-CCP emerges as a robust and practical method, offering an effective trade-off between accuracy, interpretability, and computational efficiency. Future work will focus on evaluating the method using larger and real-world datasets, and further improving its performance.

\begin{credits}
\subsubsection{\ackname} Marcos Eduardo Valle acknowledges financial support from the National Council for Scientific and Technological Development (CNPq), Brazil, under grant no 315820/2021-7, and the São Paulo Research Foundation (FAPESP), Brazil, under grant no 2023/03368-0.

\end{credits}
%
%
%
\bibliographystyle{splncs04} 
\bibliography{references}
\end{document}